# Neural Machine Translation of Logographic Languages Using Sub-character Level Information

**Longtu Zhang**
Tokyo Metropolitan University
6-6 Asahigaoka, Hino,
Tokyo 191-0065, Japan
zhang-longtu@ed.tmu.ac.jp

**Mamoru Komachi**
Tokyo Metropolitan University
6-6 Asahigaoka, Hino,
Tokyo 191-0065, Japan
komachi@tmu.ac.jp

## Abstract

Recent neural machine translation (NMT) systems have been greatly improved by encoder-decoder models with attention mechanisms and sub-word units. However, important differences between languages with logographic and alphabetic writing systems have long been overlooked. This study focuses on these differences and uses a simple approach to improve the performance of NMT systems utilizing decomposed sub-character level information for logographic languages. Our results indicate that our approach not only improves the translation capabilities of NMT systems between Chinese and English, but also further improves NMT systems between Chinese and Japanese, because it utilizes the shared information brought by similar sub-character units.

## 1 Introduction

Neural machine translation (Cho et al., 2014) (NMT) systems based on sequence-to-sequence models (Sutskever et al., 2014) have recently become the de facto standard architecture. The models use attention mechanisms (Bahdanau et al., 2015; Luong et al., 2015) to keep records of all encoding results, and can focus on particular parts of these results during decoding, so that the model can produce longer and more accurate translations. Sub-word units are another technique first introduced by Sennrich's (2016) application of the byte pair encoding (BPE) algorithm, and are used to break up words in both source and target sentences into sequences of smaller units, learned without supervision. This alleviates the risk of producing <unk> symbols when the model encounters infrequent "unknown" words, also known as the out-of-vocabulary (OOV) problem. Moreover, sub-word units, which can be viewed as learned stems and affixes, can help the NMT model better encode the source sentence and decode the target sentence, particularly when the source and target languages share some similarities.

Almost all of the methods used to improve NMT systems were developed for alphabetic languages such as English, French, and German as either the source or target language, or both. An alphabetic language typically uses an alphabet: a small set of letters (basic writing symbols) that each roughly represents a phoneme in the spoken language. Words are composed by ordered letters, and sentences are composed by space-segmented ordered words. However, in other major writing systems—namely, logographic (or character-based) languages such as Chinese, Japanese, and traditional Korean—strokes are used to construct ideographs; ideographs are used to construct characters, which are the basic units for meaningful words. Words can then further compose sentences. In alphabetic languages, sub-word units are easy to identify, whereas in logographic languages, a similar effect can be achieved only if sub-character level information is taken into consideration.[1]

Having noticed this significant difference between these two writing systems, Shi et al. (2015), Liu et al. (2017), Peng et al. (2017), and Cao et al. (2017) used stroke-level information for logographic languages when constructing word embeddings; Toyama et al. (2017) used visual information for strokes and Japanese Kanji

[1]Taking the ASPEC corpus as an example, the average word lengths are roughly 1.5 characters (Chinese words, tokenized by Jieba tokenizer), 1.7 characters (Japanese words, tokenized by MeCab tokenizer), and 5.7 characters (English words, tokenized by Moses tokenizer), respectively. Therefore, when a sub-word model of similar vocabulary size is applied directly, English sub-words usually contain several letters, which are more effective in facilitating NMT, whereas Chinese and Japanese sub-words are largely just characters.

radicals in a text classification task.[2]

Some studies have performed NMT tasks using various sub-word "equivalents". For instance, Du and Way (2017) trained factored NMT models using "Pinyin"[3] sequences on the source side. Unfortunately, they did not apply a BPE algorithm during training, and their model also cannot perform factored decoding. Wang et al. (2017) directly applied a BPE algorithm to character sequences before building NMT models. However, they did not take advantage of sub-character level information during the training of sub-word and NMT models. Kuang and Han (2018) also attempted to use a factored encoder for Chinese NMT systems using radical data. It is worth noting that although the idea of using ideographs and strokes in NLP tasks (particularly in NMT tasks) is not new, no previous NMT research has focused on the decoding process. If it is also possible to construct an ideograph/stroke decoder, we can further investigate translations between logographic languages. Additionally, no NMT research has previously used stroke data.

To summarize, there are three potential information gaps associated with current studies on NMT systems for logographic languages using sub-character level data: 1) no research has been performed on the decoding process; 2) no studies have trained models using sub-character level sub-words; and 3) no studies have attempted to build NMT models for logographic language pairs, despite their sharing many similarities. This study investigates whether sub-character information can facilitate both encoding and decoding in NMT systems and between logographic language pairs, and aims to determine the best sub-character unit granularity for each setting.

The main contributions of this study are threefold:

1. We create a sub-character database of Chinese character-based languages, and conduct MT experiments using various types of sub-character NMT models.

2. We facilitate the encoding or decoding process by using sub-character sequences on either the source or target side of the NMT system. This will improve translation performance; if sub-character information is shared between the encoder and decoder, it will further benefit the NMT system.

3. Specifically, Chinese ideograph[4] data and Japanese stroke data are the best choices for relevant NMT tasks.

## 2 Background

### 2.1 NMT with Attention Mechanisms and Sub-word Units

In this study, we applied a sequence-to-sequence model with an attention mechanism (Bahdanau et al., 2015). The basic recurrent unit is the "long short-term memory" (Hochreiter and Schmidhuber, 1997) unit. Because of the nature of the sequence-to-sequence model, the vocabulary size must be limited for the computational efficiency of the Softmax function. In such cases, the decoder outputs an `<unk>` symbol for any word that is not in the vocabulary, which will harm the translation quality. This is called the out-of-vocabulary (OOV) problem.

Sub-word unit algorithms (such as BPE algorithms) first break up a sentence into the smallest possible units. Then, two adjacent units at a time are merged according to some standard (e.g., the co-occurrence frequency). Finally, after n steps, the algorithm collects the merged units as "sub-word" units. By using sub-word units, it is possible to represent a large number of words with a small vocabulary. Originally, sub-word units were only applied to unknown words (Sennrich et al., 2016). However, in the recent GNMT (Wu et al., 2016) and transformer systems (Vaswani et al., 2017), all words are broken up into sub-word units to better represent the shared information.

For alphabetic languages, researchers have indicated that sub-word units are useful for solving OOV problems, and that shared information can further improve translation quality. The Sentencepiece project[5] compared several combinations of word-pieces (Kudo, 2018) and BPE sub-word models in English/Japanese NMT tasks. The sub-word units were trained on character (Japanese Kanji and Hiragana/Katakana) sequences. Similarly, Wang et al. (2017) attempted to compare the effects of different segmentation methods on NMT tasks, including "BPE" units trained on Chinese character sequences.

[2]To be more precise, there is another so-called syllabic writing system, which uses individual symbols to represent symbols rather than phonemes. Japanese hiragana and katakana are actually syllabic symbols rather than ideographs. In this paper, we focus only on the logographic part.

[3]An official Romanization system for standard Chinese in mainland China. Pinyin includes both letters and diacritics, which represent phonemic and tonal information, respectively.

[4]We use the term "logographic" to refer to writing systems such as Chinese characters and Japanese Kanji, and "ideograph" to refer to the character components.

[5]https://github.com/google/sentencepiece

| Character | Semantic ideograph | Phonetic ideograph | Pinyin |
|---|---|---|---|
| 驰 run | 马 horse | 也 | chн |
| 池 pool | 水(氵) water | 也 | chн |
| 施 impose | 方 direction | 也 | sh |
| 弛 loosen | 弓 bow | 也 | chн |
| 地 land | 土 soil | 也 | dм |
| 驱 drive | 马 horse | 区 | q |

Table 1: Examples of decomposed ideographs of Chinese characters. The composing ideographs of different functionality might be shared across different characters.

### 2.2 Sub-character Units in NLP

In alphabetic languages, the smallest unit for sub-word unit training is the letter; in character-based languages, the smallest units should be sub-character units, such as ideographs or strokes. Because sub-character units are shared across different characters and have similar meanings, it is possible to build a significantly smaller vocabulary to cover a large amount of training data. This has been researched quite extensively within tasks such as word embeddings, as mentioned previously.

As we can see from the examples in Table 1, there are several independent Chinese characters. Each character can be split into at least two ideographs: a semantic ideograph and a phonetic ideograph.[6] More importantly, the same ideograph can be shared by different characters denoting similar meanings. For example, the first five characters (驰, 池, 施, 弛 and 地) have similar pronunciation (and they are written similarly in Pinyin) because they share the same phonetic ideograph "也". Similarly, semantic ideographs can be shared across characters and denote a similar semantic meaning. For example, the first character "驰" and the last character "驱" share same semantic ideograph "马" (meaning "horse"); and their semantic meanings are closely related ("run" and "drive", respectively). A few ideographs can also be treated as standalone characters.

[6]Semantic ideographs denote the meaning of a character, whereas phonetic ideographs denote the pronunciation.

| Word | Meaning | Ideographs |
|---|---|---|
| 树木 | Wood | 木对木 |
| 森林 | Forest | 木木木木木 |

Table 2: Examples of multi-character words in Chinese and their ideograph sequences.

To the best of our knowledge, however, no research has been performed on logographic language NMT beyond character-level data, except in the work of Du and Way (2017), who attempted to use Pinyin sequences instead of character sequences in Chinese–English NMT tasks. Considering the fact that there are a large number of homophones and homonyms in Chinese languages, it was difficult for this method to be used to reconstruct characters in the decoding step.

## 3 NMT Using Sub-character Level Units

### 3.1 Ideograph Information

When building NMT vocabulary, the use of sub-characters (instead of words, characters, and character level sub-words) can greatly condense vocabulary size. For example, a vocabulary can be decreased from 6,000 to 10,000 character types[7] to hundreds [8] of ideographs. Table 2 presents two Chinese words composed of four different characters that have very close meanings. Character-based NMT models treat these characters separately as one-hot vectors. In contrast, if the two words are broken down into ideograph sequences, they overlap significantly. Then, only two ideographs are needed to compose the vocabulary of the two words. The computational load will be reduced, and the chances of training neurons responsible for low-frequency vocabularies will increase.

Moreover, sub-character units can serve as building blocks for constructing characters that are not present in the training data, because all CJK characters are designed to be composed of a limited number of ideographs in UNICODE standards.

[7]According to the ASPEC corpus.

[8]214 as defined in UNICODE 10.0 standard and 517 as defined in CNS11643 charset.

### 3.2 Stroke Information

All ideographs can be further decomposed into strokes, which are smaller units and have an even smaller number of types. Therefore, we also propose training our model on stroke sequences. There are five basic stroke types for Chinese characters and Japanese Kanji: "horizontal" (㇐), "vertical" (㇑), "right falling" (㇏), "left falling" (㇒), and "break" (㇕). Each stroke type can be further sub-categorized into several stroke variations. For example, left falling strokes contain both long and short left fallings (㇓ and ㇒), while a break contains many more variations, such as ㇗, ㇈, ㇆, and ㇉ (details can be found in Appendix A).

In practice, the CNS11643 charset[9] is used to transform each character into a stroke sequence, where unfortunately only "stroke-type" information is available. In this study, we manually transcribed all ideographs into stroke sequences using 33 pre-defined strokes.

### 3.3 Character Decomposition

The CNS11643 charset is used to facilitate character decomposition, where Chinese, Japanese, and Korean characters are merged into a single character type based on similarities in their forms and meanings. This is potentially beneficial; for example, if Chinese and Japanese vocabularies are built, they will authentically share some common types. There are 517 so-called "components" (i.e., ideographs) pre-defined in CNS11643. This ensures that all characters can be divided into certain sequences of components. For example, the character "可" can be split into "丁" and "口"; and the character "君" can be split into "尹" and "口". Details can be found on the CNS11643 website[10]. Using this ideograph decomposition information, all Chinese and Japanese sentence data can be transformed into new ideograph sequences; then, using the manually transcribed stroke decomposition data introduced in Section 3.2, we can also obtain new stroke sequences.

Note that although there are no clear indications of how the components/strokes are structured together, the sequence potentially contains structural information, because the writing of characters always follows a certain order, such as "up-down", "outside-in", etc. We also note that UNICODE 10.0 has introduced symbols indicating sub-character structures (Ideographic Description Characters), which provide a clearer indication of character compositions. We will make further use of this information in future studies.

| Language | Word |
|---|---|
| JP-character | 風 景 |
| JP-ideograph | 几一虫 日亠口小_1 |
| JP-stroke | 丿㇈一丨㇕一丨㇒丶<br>丨㇕一一丶一丨㇕一亅㇒丶_1 |
| CN-character | 风 景 |
| CN-ideograph | 几乂 日亠口小_1 |
| CN-stroke | 丿㇈㇒丶<br>丨㇕一一丶一丨㇕一亅㇒丶_1 |
| EN | landscape |

Table 3: The Japanese word 風景 and Chinese word 风景 both mean "landscape" in English, and they only differ in the middle part of the first character. Note that there are "_1" tags at the ends of some decomposed sequences to distinguish between possible duplications.

To ensure that there are no duplicated ideograph and stroke sequences for different characters and multi-character words, we post-tag the sequences on the duplicated ones using "_1", "_2", etc. Table 3 shows an example of character decomposition in Chinese and Japanese[11].

## 4 Experiments on Chinese–Japanese–English Translation

To answer our research questions, we set up a series of experiments to compare NMT models of logographic languages trained on word sequences, character-level sub-word unit sequences, and ideograph- and stroke-level sub-word unit sequences.

We performed two lines of experiments:

1. We trained NMT models between logographic language and alphabetic language combinations, i.e., Japanese/Chinese and English. In each model, we varied the data granularity for the logographic language, using "character level" or "sub-character level" (ideograph level and stroke level) granularities. We used the character level NMT models as our baselines, and investigated whether the sub-character level NMT models could outperform the baseline models.

[9] The CNS11643 charset is published and maintained by the Taiwan government. http://www.cns11643.gov.tw/AIDB/welcome_en.do

[10] http://www.cns11643.gov.tw/search.jsp?ID=13

[11] For example, the ideograph and stroke sequences for character 景 are the same as those for character 晾 (meaning "to dry in the sun"). However, these two characters have different architectures ("top-down" vs. "left-right"). Post-tags are thus appended in order to distinguish them. Similarly, characters 风 and 凩 have the same ideograph and stroke sequences, and thus must be post-tagged.

2. We trained NMT models between combinations of two logographic languages, i.e., Chinese and Japanese. Similarly, we used data sets with different granularities: 1) Models lacking sub-character level data. 2) Models having sub-character level data on both sides (to confirm the results of the previous experiment). For the experiments, the models will have both source and target sides. The models will use sub-character level data with/without shared vocabularies (namely, ideograph models, stroke models, ideograph-stroke models, stroke-ideograph models, and ideograph/stroke models with shared vocabularies). 3) Pinyin baselines according to (Du and Way, 2017), where both Pinyin word sequences with tones and character sequences with Pinyin factors are used with the encoder.

### 4.1 Dataset

We trained our baselines and experiments using Chinese, Japanese, and English. The Asian Scientific Paper Excerpt Corpus (ASPEC (Nakazawa et al., 2016)) and Casia2015[12] corpus were used for this purpose.

ASPEC contains a Japanese–English paper abstract corpus of 3 million parallel sentences (ASPEC-JE) and a Japanese–Chinese paper excerpt corpus of 680,000 parallel sentences (ASPEC-JC). We used the first million confidently aligned parallel sentences in ASPEC-JE and all of the ASPEC-JC data to cover Japanese–English and Japanese–Chinese language pairs. The Casia2015 corpus contains approximately 1 million parallel Chinese–English sentences. All data in the Casia2015 corpus were used to cover Chinese–English language pairs. During training, the maximum length hyperparameter was adjusted to ensure 90% coverage of the training data. For development and testing, the ASPEC corpus has an official split between the development set and test set; however, because the Casia2015 corpus is not similarly split, we made random selections from the development set and test set of 1,000 sentences each.

### 4.2 Settings

Different pre-tokenization methods were applied to the data in three languages (if applicable). A Moses tokenizer was applied to the English data; a Jieba[13] tokenizer using the default dictionary was applied to the Chinese data; and a MeCab[14] tokenizer using the IPA dictionary was applied to the Japanese data. For the Pinyin baseline, the pypinyin[15] Python library was used to transcribe the Chinese character sequence into a Pinyin sequence.

In both of the experiment lines discussed above, data at the "word", "character", "ideograph", and "stroke" levels were used in combinations. For "word" level data, only dictionary-based segmentation was applied; for the other three levels of data, the byte pair encoding (BPE) models were trained and applied, with a vocabulary size of 8,000. In the second line of experiments, where both the source and target sides were logographic languages, we added "character" level data without BPE ("char") for comparison. Additionally, shared vocabularies were applied when both the source and target had the same data granularity level (meaning that both the source and target side would have the same vocabulary)[16].

A basic RNNsearch model (Bahdanau et al., 2015) with two layers of long short-term memory (LSTM) units was used. The hidden size was 512. A normalized Bahdanau attention mechanism was applied at the output layer of the decoder. We developed our model based on TensorFlow[17] and its neural machine translation tutorial[18].

The model was trained on a single GeForce GTX TITAN X GPU. During training, the SGD optimizer was used, and the learning rate was set at 1.0. The size of the training batch was set to 128, and the total global training step was 250,000. We also decayed the learning rate as the training progressed: after two-thirds of the training steps, we set the learning rate to be four times smaller until the end of training. Additionally, we set the drop-out rate to 0.2 during training.

[12] http://nlp.nju.edu.cn/cwmt-wmt/, provided by the Institute of Automation, Chinese Academy of Sciences.

[13] https://github.com/fxsjy/jieba

[14] http://taku910.github.io/mecab/

[15] https://github.com/mozillazg/python-pinyin

[16] The shared vocabulary can be trained by a BPE model on a concatenated corpus of source and target sentences.

[17] https://github.com/tensorflow

[18] https://github.com/tensorflow/nmt

| English-Japanese NMT | | BLEU |
|---|---|---|
| EN_word | JP_word | 36.1 |
| EN_word | JP_character | 38.3 |
| EN_word | JP_ideograph | 40.3* |
| EN_word | JP_stroke | **41.3*** |
| **Japanese-English NMT** | | **BLEU** |
| JP_word | EN_word | 25.5 |
| JP_character | EN_word | 26.3 |
| JP_ideograph | EN_word | 26.8* |
| JP_stroke | EN_word | **27.0*** |
| **English-Chinese NMT** | | **BLEU** |
| EN_word | CN_word | 11.8 |
| EN_word | CN_character | 10.3 |
| EN_word | CN_ideograph | **14.6*** |
| EN_word | CN_stroke | 14.1* |
| **Chinese-English NMT** | | **BLEU** |
| CN_word | EN_word | 14.7 |
| CN_character | EN_word | 14.5 |
| CN_ideograph | EN_word | **15.6*** |
| CN_stroke | EN_word | 15.5* |

Table 4: Experimental results (BLEU scores) of NMT systems for Japanese/English and Chinese/English language pairs. All the scores are statistically significant at $p = 0.0001$ (marked by $*$).

BLEU was used as the evaluation metric in our experiments. For Chinese and Japanese data, a KyTea tokenization was applied before we applied BLEU, following the WAT (Workshop on Asian Translation) leaderboard standard. To validate the significance of our results, we ran bootstrap re-sampling (Koehn, 2004) for all results using Travatar (Neubig, 2013) at a significance level of $p = 0.0001$.

## 4.3 Results

### 4.3.1 NMT of Logographic and Alphabetic Language Pairs

Table 4 shows the experimental results for the Japanese/English and Chinese/English language pairs in both translation directions. Generally, for each of the experiment settings, the models using ideograph and stroke data outperformed the baseline systems, regardless of the language pair or translation direction. However, for the Japanese/English language pair, the stroke sequence models performed better. For the Chinese/English language pairs, the ideograph sequence models worked better. The reason for these differences will be discussed in detail in Section 5.

### 4.3.2 NMT of Logographic Language Pairs

Table 5 shows the results for all baselines and proposed models. Among the character-level baselines, the "char" models and "bpe" models outperformed the "word" models in both translation directions. When we applied a shared vocabulary to the "bpe" models, the models achieved the best BLEU scores in both translation directions. These character-level baselines conform with previous studies indicating that sub-word units improve the performance of NMT systems, and that whenever both the source and target side data have similarities in their writing systems, shared vocabularies will further enhance performance.

Sub-character level models aim to replicate similar results to those presented in Section 4.3.1, because only one side of these models uses sub-character level data. For Japanese–Chinese translation directions, half of the models showed a significant improvement over the baselines, whereas for Chinese–Japanese translation directions, five out of six models showed significant improvements.

When both the source and target side used the same sub-character level data (either ideograph or stroke data), the experimental results also showed significant improvement over character baselines. Additionally, the ideograph models outperformed stroke models. When shared vocabularies were applied to the models, the ideograph models exhibited slight performance improvements ($0.1 \sim 0.4$ BLEU point), and the stroke models exhibited dramatically decreased performance ($0.9 \sim 1.1$ BLEU points). However, no model here outperformed the sub-character baselines.

To further exploit the power of sub-character units, the last models having different levels of sub-character units on the source and target side were trained. The results conform with what we found in Section 4.3.1: the models using Chinese ideograph data and Japanese stroke data exhibited the best performance, regardless of whether they were applied at the source or target side. For Japanese–Chinese translations, the best BLEU score was 33.8, which was produced by the Japanese-stroke and Chinese-ideograph model; for Chinese–Japanese translation, the best BLEU

| **JP-CN NMT** | **CN_word** | **CN_char** | **CN_bpe** | **CN_ideograph** | **CN_stroke** |
|---|---|---|---|---|---|
| **JP_word** | 29.6 | - | - | 30.8 | 30.3 |
| **JP_char** | - | 31.6 | - | 32.0* | 32.1* |
| **JP_bpe** | - | - | 31.5 (31.7) | 31.6 | 31.7 |
| **JP_ideograph** | 30.4 | 33.1* | 33.3* | 32.0* (32.4*) | 33.4* |
| **JP_stroke** | 30.3 | 33.4* | 32.6* | **33.8*** | 32.1* (31.2) |
| **CN-JP NMT** | **JP_word** | **JP_char** | **JP_bpe** | **JP_ideograph** | **JP_stroke** |
| **CN_word** | 40.0 (*40.0*) | - | - | 40.5 | 40.1 |
| **CN_char** | 42.1 (*40.4*) | 41.7 | - | 43.1* | 42.2* |
| **CN_bpe** | 42.1 | - | 42.0 (42.3) | 43.1* | 42.2* |
| **CN_ideograph** | 43.2* | 43.5* | 43.0* | 42.6* (42.7*) | **43.9*** |
| **CN_stroke** | 43.0* | 43.3* | 42.5* | 42.9* | 42.2* (41.1) |

Table 5: Experimental results (BLEU scores) for Japanese/Chinese NMT systems. The row headers and column headers indicate which source and target data were used in the training. In particular, "word" and "char" are character level data without BPE segmentation, while "bpe" (character level), "ideograph", and "stroke" (sub-character level) are data with BPE segmentation. The scores in parentheses indicate the models that had a shared vocabulary, whenever applicable. The italic numbers represent the two Pinyin baselines used for comparative purposes, namely the "WdPyT" model, which uses Pinyin words with tones as the source data, and the "factored-NMT" model, which uses Pinyin characters as factors (Du and Way, 2017). Note that these two baselines can only have Chinese data on the encoder side. The $*$ superscripts indicate that a score is significantly better than the best baseline result.

score was 43.9, which was produced by the Chinese-ideograph and Japanese-stroke model.

# 5 Discussions

## 5.1 Translation Examples

Table 6 shows some of the translation examples. There is a rare proper noun "松下電器 (Matsushita Electric)" (OOV) in the source sentence. The word baseline model cannot decode this; therefore, an <unk> symbol is produced. The character baseline model avoids the OOV problem. However, the underlined parts in both baseline translations seem to be word-for-word translations from the Japanese source sentence ("松下 電器 グループ で は"), which become a prepositional phrase in Chinese ("在 松下 电器 集团 中 (in Matsushita Electric Group)"). This makes the translation ungrammatical because there will be no noun phrase as the subject in the sentence. Our best model (i.e., sub-character based NMT model using Japanese stroke data and Chinese ideograph data) can solve these two problems by better encoding the source sentence and can produce translations both without OOV and with a noun phrase as the sentence subject.

## 5.2 Strokes vs. Ideographs

The experimental results show that in NMT models, different logographic languages appear to prefer sub-character units with different granularities. A very clear tendency that was observed consistently in both experiments was that ideographs worked better for Chinese and strokes worked better for Japanese. This difference might be because of the differences in the writing systems. In addition to Kanji (Chinese characters), Japanese uses Hiragana and Katakana, which are standalone alphabets.

Moreover, as described in Section 4, stroke models tended to perform more poorly than ideograph models. This probably occurred because to achieve a fair comparison between all baseline models and proposed models, the same hyper-parameter configurations were used. For example, the embedding dimensions for all vocabularies were set to 300. This might be appropriate for vocabularies of character-based data and ideograph data having vocabulary sizes larger than 500. However, the stroke data only has a vocabulary size of approximately 30, which is too disproportional. This phenomenon might also account for the decrease in BLEU scores when shared vocabularies were applied to stroke models.

## 5.3 The Encoding and Decoding Process

In comparison with character level data, sub-character level data (such as ideographs and strokes) can be used to generate much smaller and more concentrated vocabularies. This is helpful during both the encoding and decoding pro-cesses. Vocabularies constructed using character-level data are known to be very skewed, containing both very frequent words and very rare words. As a result, during training, the neurons responsible for high-frequency words might be updated many times, while the neurons responsible for low-frequency words might be updated only a very limited number of times. This will potentially harm translation performance for low-frequency words.

| Model | Sentence |
|---|---|
| Source | 松下 電器 グループ で は ， 経営 理念 の 基本 と し て 1991 年 に 「 環境 宣言 」 を 制定 し た 。 |
| Reference | 作为 经营 理念 ， 松下电气集团 于 1991年 制定 了 《 环境 宣言 》 。 |
| Baseline (Word) | 在 <unk> 集团 中 ， 1991年 制定 了 “ 环境 宣言 ” 作为 经营 理念 的 基础 。 |
| Baseline (Char) | 在 松下电器 集团 中 ， 作为 经营 理念 的 基础 ， 1991年 制定 了 《 环境 宣言 》 。 |
| Best Model (JP-stroke-CN-ideograph) | 松下电器集团 ， 作为 经营 理念 的 基础 ， 1991年 制定 了 “ 环境 宣言 ” 。 |
| English Translation | The Matsushita Electric Group enacted the ”Environmental Declaration” as the basis of its business philosophy in 1991. |

Table 6: Translation examples of Japanese-Chinese NMT systems. Note that “松下电器” as a proper noun, could be handled properly in sub-character based translation systems.

However, this problem can be alleviated by applying sub-character units. Because ideographs and strokes are repeatedly shared by different characters, no items occur with very low frequencies. More instances can be found in the training data, even for the least frequent sub-character items. Therefore, the translation performance for low-frequency items could be much better.

## 6 Conclusions and Future Work

This study was the first attempt to use sub-character units in NMT models. Our results not only confirmed the positive effects of using ideograph and stroke sequences in NMT tasks, but also indicated that different logographic languages actually preferred different sub-character granularities (namely, ideograph for Chinese and stroke for Japanese). Finally, this paper presented a simple method for extending the available corpus from the character level to the sub-character level. During this process, we maintained a one-to-one relationship between the original characters and transformed sub-character sequences. As a result, this simple and straightforward method achieved consistently better results for NMT systems used to translate logographic languages, and could be easily applied to similar scenarios.

Many questions remain to be answered in future work. The first question involves the relative benefits of ideograph data and stroke data, and the effects of shared vocabularies. We have not yet explained why there are considerable differences in performance. In particular, for NMT models in which both sides have stroke data, why does performance drop dramatically when shared vocabularies are applied? We discussed the possible reasons for this phenomenon in Section 5.2; however, further investigation is needed.

Another important issue is as follows: when characters are transformed into ideographs and strokes, no structural information is considered. This causes repetitions in data, and we must add tags at the end of each sequence to differentiate them. A better way to solve this problem would be to have structural information directly encoded in the data.

## Acknowledgments

This work was partially supported by JSPS Grant-in-Aid for Young Scientists (B) Grant Number JP16K16117.

## Appendix A Strokes in CNS11643 Charset

| Type | Strokes |
| --- | --- |
| Horizontal |  |
| Vertical |  |
| Left-falling | 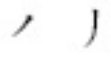 |
| Right-falling | 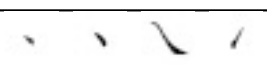 |
| Break | 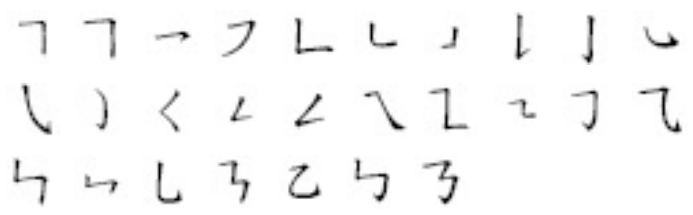 |